%% file: acl2023.tex
\pdfoutput=1

\documentclass[11pt]{article}

\usepackage[usenames,dvipsnames]{xcolor}
\usepackage[square,numbers]{natbib}
\usepackage{enumitem, graphicx}
\usepackage{amssymb}
\usepackage{pifont}
\newcommand{\xmark}{\ding{55}}%

\usepackage{ACL2023}

\usepackage{times}
\usepackage{latexsym}

\usepackage[T1]{fontenc}

\usepackage[utf8]{inputenc}

\usepackage{microtype}

\usepackage{inconsolata}

\title{ChatGPT is fun, but it is not funny! \\Humor is still challenging Large Language Models}

\author{Sophie Jentzsch$^{1}$ \and Kristian Kersting$^{2,3,4}$ \\
  $^1$Institute for Software Technology, German Aerospace Center (DLR), Cologne, Germany\\
  $^2$Computer Science Department, Technical University Darmstadt, Darmstadt, Germany\\ 
  $^3$Centre for Cognitive Science, TU Darmstadt \\
  $^4$Hessian Center for AI (hessian.AI) \\
  sophie.jentzsch@dlr.de, kersting@cs.tu-darmstadt.de
  }

\begin{document}
\maketitle
\begin{abstract}
Humor is a central aspect of human communication that has not been solved for artificial agents so far. 
Large language models (LLMs) are increasingly able to capture implicit and contextual information. Especially, OpenAI's ChatGPT recently gained immense public attention. %
The GPT3-based model almost seems to communicate on a human level and can even tell jokes. 
Humor is an essential component of human communication. 
But is ChatGPT really funny? 

We put ChatGPT's sense of humor to the test. 
In a series of exploratory experiments around jokes, i.e., generation, explanation, and detection, we seek to understand 
ChatGPT's capability to grasp and reproduce human humor. Since the model itself is not accessible, we applied prompt-based experiments. %

Our empirical evidence indicates that jokes are not hard-coded but mostly also not newly generated by the model. Over 90\% of $1008$ generated jokes were the same 25 Jokes. The system accurately explains valid jokes but also comes up with fictional explanations for invalid jokes. Joke-typical characteristics 
can mislead ChatGPT in the classification of jokes. ChatGPT has not solved computational humor yet but it can be a big leap toward ``funny'' machines. 
\end{abstract}

\newcommand{\p}{\textcolor{ForestGreen}{[P]}}
\newcommand{\h}{\textcolor{Dandelion}{[H]}}
\newcommand{\n}{\textcolor{BrickRed}{[N]}}

\section{Introduction}
For humans, humor plays a central role in forming relationships and can enhance performance and motivation~\citep{savage2017humor}. It is a powerful instrument to affect emotion and guide attention~\citep{binsted2006computational}. Thus, a computational sense of humor holds the potential to massively boost human-computer interaction (HCI). 
Unfortunately, although computational humor is a longstanding research domain~\citep{winters2021computers}, 
the developed machines are far from "funny." This problem is even considered to be AI-complete~\citep{veale2021your}. 

Continuous advances and recent breakthroughs in machine learning (ML), however, lead to an ever-increasing panoply of applications [e.g.,~\citealp{rombach2021highresolution},~\citealp[]{frosio2023artificial}] and likewise open new opportunities for natural language processing (NLP). 
Transformer-based large language models (LLMs) increasingly capture and reflect implicit information, such as stereotypes~\citep{jentzsch2022gender}, moral~\citep{jentzsch2019semantics}, and humor~\citep{hasan2021humor,weller2019humor}. Humor is often implicit and carried by subtle details. Thus these novel qualities of LLMs give reason to hope for new advances in artificial humor. 

\begin{figure}[t]
\vskip 0.2in
\begin{center}
\centerline{\includegraphics[width=\columnwidth]{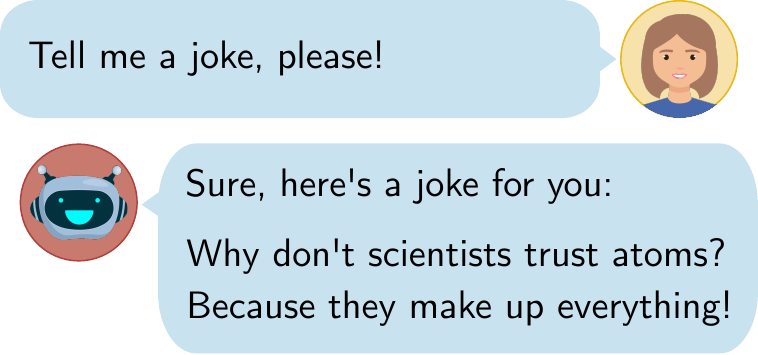}}
\caption{Exemplary illustration of a conversation between a human user and an artificial chatbot. The joke is a true response to the presented prompt by ChatGPT.\label{fig:screenshot}}  
\end{center}
\vskip -0.2in
\end{figure}

Most recently, OpenAI's ChatGPT gained immense attention for its unprecedented capabilities. Users can interact with the model via public chat API in a conversation-like course. 
The system is able to answer a huge variety of questions while taking the previous contextual conversation into consideration. In fact, it can even tell jokes, as displayed in Fig.~\ref{fig:screenshot}. ChatGPT is fun and almost appears to interact on a human-like level. 
Yet, when interacting with the model, users may quickly get a glimpse of its limitations. Although ChatGPT generates text in almost error-free English, grammatical glitches and content-related mistakes occur. 
In the preceding exploration, we noticed that ChatGPT is likely to repeat the exact same jokes frequently. 
Moreover, the provided jokes were strikingly correct and sophisticated. These observations led to the hypothesis that output jokes are not originally generated by the model. Instead, they seem to be reproduced from the training data or maybe even hard coded in a predefined list. 

As the system's inner workings are not shared, we conducted a series of structured prompt-based experiments to learn about the system's behavior and allow for inference regarding the generation process of ChatGPT's output. Specifically, we aim to understand to what extent ChatGPT is able to capture human humor by conducting a systematic prompt-based analysis. 

The remainder of this paper is structured as follows: First, Sec.~\ref{S:relatedwork} discusses related research. The main contribution assembles three experimental conditions: Joke generation, joke explanation, and joke detection. While the overall approach is outlined in Sec.~\ref{S:method}, the detailed conduction is explained together with the results of each experiment in Sec.~\ref{S:joke_gen}, Sec.~\ref{S:joke_explanation}, and Sec.~\ref{S:joke_classification}, respectively. We close by discussing implications and further research in Sec.~\ref{S:discussion} and conclude our work in Sec.~\ref{S:conclusion}.

\textbf{Terminology Disclaimer.}
AI-related terminology tends to make use of analogies to human characteristics, such as \emph{neural networks}, or the term artificial \emph{intelligence} itself. 
Likewise, we use human-related terms in the discussion around conversational agents, which are supposed to mimic human behavior as accurately as possible, e.g., ChatGPT "\textit{understands}" or "\textit{explains}." 
Although we believe that these analogies describe the system's behavior and its inner workings very well, they might be misleading. We would like to emphasize that the discussed AI models are not on a human level and, at best, constitute a simulation of the human mind. Whether AI can potentially ever \emph{think} or \emph{understand} in a conscious way is a philosophical question that is out of the scope of this investigation. 

\section{Related Work} \label{S:relatedwork}
Our work considers the intersection of two lines of research, namely LLMs and computational humor.

\textbf{Large Language Models.} NLP witnessed several leaps in the recent past. %
Only a few years ago, the introduction of transformer architectures in neural networks~\citep{vaswani2017attention} enabled the development of contextualized models, such as BERT~\citep{kenton2019bert} or GPT~\citep{brown2020language}. %
These so-called large language models (LLMs) capture relations in the natural data and even reflect implicitly-included information, which can be both a risk~\citep{wagner2022gender} and a chance~\citep{schramowski2022large}. Either way, however, it is a prerequisite to solving computational humor. 

OpenAI\footnote{OpenAI, \url{https://openai.com/}} recently released ChatGPT, a GPT3.5-based LLM that is able to interact with users conversationally through a chat-like API. 
It is fine-tuned with \textit{Reinforcement Learning from Human Feedback (RLHF)}~\citep{ouyangtraining} in three essential steps, including large-scale manual labeling and proximal policy optimization~\citep{schulman2017proximal}. 
The result of this process is a model with unprecedented capabilities: It interacts in a conversational manner, i.e., it refers to the users' prompt while also taking information from the preceding conversation into account. It is able to summarize information, generate new texts of all shapes, and even write program code. 
Nevertheless, there are still glitches and limitations, e.g., factual wrong information presented as valid. 

\textbf{Computational Humor.} 
Computational humor is a small but active research area of NLP with the main streams \textit{Humor Detection} or \textit{Classification} [e.g.,~\citealp{kayatani2021laughing},~\citealp{guo2022federated}], and \textit{Humor Generation} [e.g.,~\citealp{stock2005hahacronym}]. %
Although advances in ML and NLP transfer to computational humor, researchers highlight the great complexity of automated humor and the limitations of current endeavors~\citep{winters2021computers,wang2021dehumor}. 
Humor is one of the few capabilities that seemed to be reserved for human individuals thus far, which is why it is even considered AI-complete [\citealp{binsted2006computational},~\citealp{veale2021your}].

While decades of research in linguistics and psychology offer quite a few logic-based humor theories [e.g.,~\citealp{raskin1979semantic},~\citealp{zillmann1980misattribution}], the work in the context of artificial agents is predominantly data-driven. In these approaches, pretrained language models, such as ChatGPT, play a central role~\citep{meaney2021semeval}. With their innovative capabilities, GPT-based models have the potential to open a new chapter of computational research. 

\begin{figure*}[t]
\vskip 0.2in
\begin{center}
\centerline{\includegraphics[width=16cm]{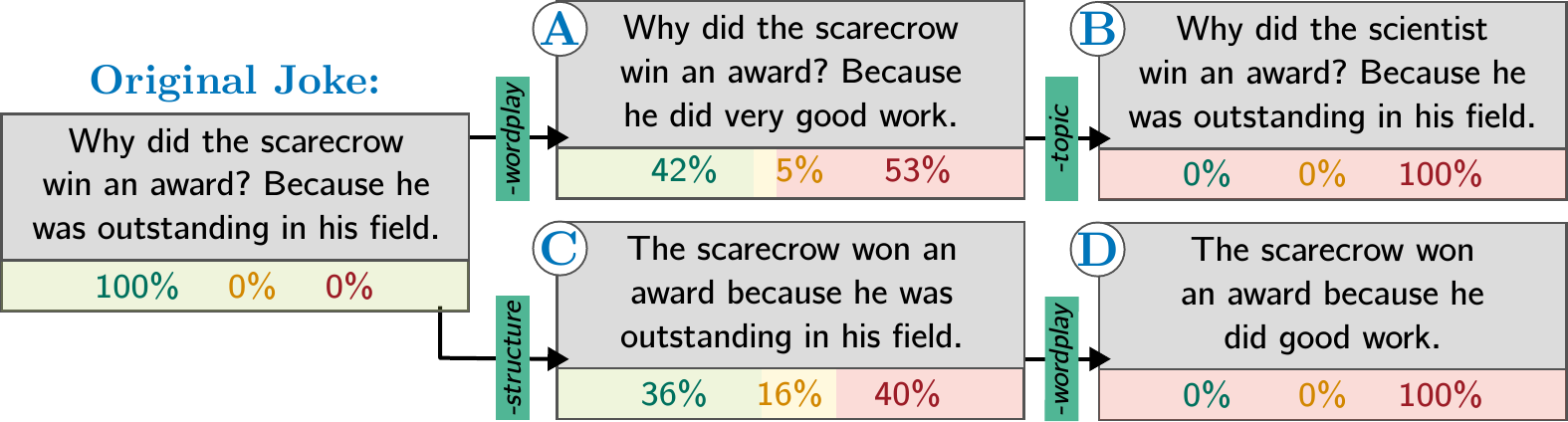}}
\caption{Modification of top jokes to create joke detection conditions. Below each condition, the percentages of samples are stated that were classified as joke (green), potentially funny (yellow), and not as a joke (red). In condition \textit{(A)~Minus Wordplay}, the comic element, and, therefore, the pun itself, was removed. For condition \textit{(B)~Minus Topic}, the joke-specific topic was additionally eliminated, e.g., by removing personifications. Condition \textit{(C)~Minus Structure} keeps the validity of the joke intact but changes the typical q-a-structure to a single-sentence-sample. From that, the comic element was additionally removed to create condition \textit{(D)~Minus Wordplay.}}
\label{Fig:modification}
\end{center}
\vskip -0.2in
\end{figure*}

\section{Method}\label{S:method}
The presented experiments are grouped into three individual tasks, which are introduced in Sec.~\ref{ss:meth_jokegen}, Sec.~\ref{ss_meth_jokeexpl}, and Sec.~\ref{ss_meth_jokedet}. Implementation details and extended results are made available at GitHub\footnote{Project repository: \\ \url{https://github.com/DLR-SC/JokeGPT-WASSA23}}. %
In all experiments, each prompt was conducted in a new empty conversation to avoid unwanted influence. To conduct a large number of prompts with OpenAI's ChatGPT\footnote{ChatGPT user API at \url{chat.openai.com/}, Accessed: January-March 2023 (detailed dates in the Appendix)} web service, there were certain obstacles to overcome. %
Since there was no official API available at the time, prompts were entered with the help of a wrapper. The number of permitted prompts per hour was limited to $74$. Moreover, ChatGPT was unavailable for longer periods due to exceeded capacity. 

In this work, we differentiate between \textit{originally generated} output, i.e., text composed by the model, and \textit{replicated} output, i.e., text that is memorized from training data and played back by the system in the exact wording. \textit{Modified} output is a mix of both, i.e., replicated text that is slightly altered, e.g., mixing the first half of one joke with the second half of another. %
We classify a joke as \textit{valid} if it is funny from a human perspective. 
Accordingly, an \textit{invalid} joke might be grammatically correct and even contain joke-like elements but fails to deliver a punch line. 
Naturally, 
as humor is subjective, 
these categories are always debatable. That being said, the distinction is comparably evident for the present examples, as we expound in the following chapters. 

\subsection{Joke Generation}\label{ss:meth_jokegen} To test if there is a limited number of reoccurring jokes, we analyze the deviation of output jokes. To this end, we asked ChatGPT to provide a joke a thousand times. 
We applied a predefined list of ten differently worded prompts, such as \textit{"Can you tell me a joke, please?"} or \textit{"I would love to hear a joke."} The resulting observations are described in Sec.~\ref{S:joke_gen}. We identified 25 repeating top jokes, which form the basis for the two subsequent tasks. 

\subsection{Joke Explanation}\label{ss_meth_jokeexpl} In the joke generation task, it was tested whether ChatGPT is able to generate valid jokes. However, this task does not necessarily reflect the system's ability to \emph{understand} humor, 
i.e., \emph{why} the joke might be funny from a human perspective. To see to what extent the model captures these complex inner workings of jokes, we asked ChatGPT to explain each of the generated top 25 jokes. The prompt was always \textit{"Can you explain why this joke is funny:} [joke]." The results from this second task are presented in Sec.~\ref{S:joke_explanation}.

\subsection{Joke Detection}\label{ss_meth_jokedet} In the first two tasks, we identified certain criteria that (almost) all output joke samples had in common, i.e., structure, topic, and wordplay. These criteria seemed to be central joke characteristics for ChatGPT. To examine how close these cues are connected to ChatGPT's conception of humor, we manually modified the top 25 jokes to eliminate one or multiple of the three criteria, resulting in four additional conditions for these jokes. We asked the system to classify each sample with the prompt "\textit{What kind of sentence is that:} [sample]." ChatGPT's response would then either include a categorization as a joke or not, as explained in Sec.~\ref{S:joke_classification}. %
The three defined joke characteristics were defined as follows: 
\begin{description}
\item{\bf Structure:} The generated jokes from Sec.~\ref{S:joke_gen} were in noteworthy similar semantic structure. Despite one sample, all $1008$ generated jokes were in the same question-answer format. 
\item{\bf Comic element:} In jokes, there is usually a stylistic element that creates comic momentum. ChatGPT's generated jokes exclusively contained wordplay puns, e.g., the double meaning of one word. 
\item{\bf Topic} Joke scenarios tend to be bizarre and not close to reality. Not always, but often they contain personifications of objects, i.e., protagonists can be computers or bananas. 
\end{description}

To determine the impact of these characteristics on the classification result, we compared the original top 25 jokes to samples with one or multiple of these characteristics removed. The considered jokes were modified manually as described in Fig.~\ref{Fig:modification} to create alternative samples that were semantically as similar as possible. The comprehensive sets of samples and their classification can be found in the Appendix in Sec~\ref{A_joke_detection}. The sets were created as follows.

In the first modification~\textit{A}, the wordplay was removed from the joke (\emph{minus wordplay}). To achieve that, the term(s) that form(s) the center of the pun were replaced by a wording with comparable primary meaning but without double meaning. 
As a side effect, this step removes the comic element and therefore destroys the pun. The joke would not be perceived as funny from a human perspective. %
If the jokes contained a joke-like topic, e.g., an award-winning scarecrow, this was removed in a second step (\emph{minus topic}) by replacing it with an everyday entity, e.g., a scientist, to achieve modification~\textit{B}. 
In case the original sample did not contain an unrealistic joke-specific topic, such as "\textit{Why did the man put his money in the freezer?}", it was included in set~\textit{B} and not~\textit{A}. Thus, samples of the set~\textit{A} contained a joke topic but no comic element ($N = 19$), and samples of the set~\textit{B} included none of both ($N = 25$).

Eliminating the question-answer format for modification~\textit{C}, i.e., \emph{minus structure}, was straightforwardly possible for all 25 original jokes ($N = 25$) by rewriting it in the format "[sentence one] \textit{because} [sentence two]." In this case, the pun remained intact, and the joke was similarly funny. The original topic remained unchanged. Then, the comic element, i.e., the wordplay, was again removed to form set~\textit{D} ($N = 25$). 

\section{Results}
With this design at hand, let us now turn to our empirical evidence gathered on joke generation, explanation, and detection. 

\subsection{Joke Generation}\label{S:joke_gen}
To test how rich the variety of ChatGPT's jokes is, we asked it to tell a joke a thousand times. All responses were grammatically correct. Almost all outputs contained exactly one joke. Only the prompt \textit{do you know any good jokes?} provoked multiple jokes, leading to $1008$ responded jokes in total. 
Besides that, the variation of prompts did have any noticeable effect. 

To extract the deviation of jokes in the set of responses, similar samples were grouped. %
Removing direct duplicates 
reduced the number of individual samples to $348$. 
Then, we removed opening sentences, such as \textit{"How about this one"} in the example in Fig.~\ref{fig:screenshot}, and minor formatting differences, such as extra line breaks. This resulted in a list of $128$ individual responses. %
Finally, some samples could again be grouped together, such as in Ex.1. 
\paragraph{Example 1.} The following samples are no direct duplicates, as the wording is slightly different. However, they represent the same pun and are therefore grouped together. 
\begin{itemize}[leftmargin=1cm]\itemsep0em 
    \item[(1.1)] \textit{Why did the bicycle fall over? \\
    \textbf{Because it} was two-tired. } 
    \item[(1.2)] \textit{Why didn’t the bicycle \textbf{stand up by itself}? \\
    Because it was \textbf{two tired}.}
\end{itemize}

These steps resulted in a final list of 25 top frequent jokes.

\paragraph{Top 25 Jokes.} 
The final list of the top 25 jokes covered $917$ of $1008$ samples and can be found in the Appendix in Sec.~\ref{A_joke_generation}. Jokes are presented together with their number of occurrences. These are the five most frequent jokes: 
\begin{itemize}[leftmargin=1cm]\itemsep0em 
    \item[T1.] \textit{Why did the scarecrow win an award? Because he was outstanding in his field.} (140)
    \item[T2.] \textit{Why did the tomato turn red? \\
    Because it saw the salad dressing.} (122)
    \item[T3.] \textit{Why was the math book sad? \\
     Because it had too many problems.} (121)
    \item[T4.] \textit{Why don't scientists trust atoms? \\
     Because they make up everything.} (119)
    \item[T5.] \textit{Why did the cookie go to the doctor? \\
     Because it was feeling crumbly.} (79)
\end{itemize}
The number of occurrences among these examples varies largely. While the top four jokes occurred over a hundred times each, the jokes T13 to T25 occurred less than 20 times.
All 25 puns together sum up to about ${90\%}$ of the gathered puns, but the top four examples alone make more than ${50\%}$. This observation contradicts our initial hypothesis: In the case of randomly picking from a predefined list, we would expect the occurrence of samples to be more equally distributed. Nevertheless, the small number of repeating samples indicates a limited versatility in ChatGPT's response pattern. %

All of the top 25 samples are existing jokes. They are included in many different text sources, e.g., they can immediately be found in the exact same wording in an ordinary internet search. Therefore, these examples cannot be considered original creations of ChatGPT. 

Of $1008$ samples, $909$ were identical to one of the top 25 jokes. The remaining $99$ samples, however, did not necessarily contain new content. About half of them were again modifications of the top jokes, as illustrated by the examples Ex.~2, Ex.~3, and Ex.~4. While some of the modified puns still made sense and mostly just replaced parts of the original joke with semantically similar elements, others lost their conclusiveness. Thus, although the top 25 joke samples rather appear to be replicated than originally generated, there seems to be original content in the remaining samples. 

\paragraph{Example 2.}
Item~{2.1} is the famous chicken joke - a so-called anti-joke: It creates expectations with its joke-typical build-up but omits the reliving punch line. Besides that original joke, many variations exist with the chicken in another situation but a similar-sounding anti-pun.
Item~{2.2} is such a variation and is even more represented in the set of generated samples than in the original joke. 
Items~{2.3}, {2.4}, and {2.5} are not covered by the top 25 jokes and can be considered modifications by ChatGPT, e.g., by replacing "chicken" in Item~2.2 with a semantically similar concept, i.e., "duck," to create Item~2.5.
\begin{itemize}[leftmargin=1cm]\itemsep0em 
    \item[(2.1)] \textit{Why did the chicken cross the road? \\
    To get to the other side. (7)}
    \item[(2.2)] \textit{Why did the chicken cross the playground? \\
    To get to the other slide. (33)}
    \item[(2.3)] \textit{Why did the duck cross the road? \\
    To get to the other pond. (2)}    
    \item[(2.4)] \textit{Why did the chicken wear a tuxedo? \\
    Because it was a formal occasion. (1)}
    \item[(2.5)] \textit{Why did the duck cross the playground? \\
    To get to the other slide. (1)}  
\end{itemize}
For anti-jokes, it is especially hard to tell whether a sample is valid, as they do not compute in the classical sense. 
Yet, it is safe to say that the first two items are already existing jokes, and the latter ones are rather rare or even generated by ChatGPT.

\paragraph{Example 3.} While it is debatable whether we observe that behavior in Ex.~1, Ex~2. clearly illustrates how ChatGPT mixes up elements from different valid jokes and, by that means, creates new samples. Item~3.1 is one of the top 25 jokes. The Items~3.2 and 3.3 were in the remaining $99$ not-grouped jokes. 
\begin{itemize}[leftmargin=1cm]\itemsep0em 
    \item[(3.1)] \textit{Why did the man put his money in the freezer? He wanted cold hard cash.} (13)
    \item[(3.2)] \textit{Why did the man put his watch in the blender? He wanted to make time fly.} (2)
    \item[(3.3)] \textit{Why did the man put his money in the blender? He wanted to make time fly.} (2)
\end{itemize}
Item~{3.1} and Item~{3.2} are valid replicated jokes, and Item~{3.3} is an invalid combination of those. 

\paragraph{Example 4.}
One of the most frequent jokes in the output list is the math book pun, Item~{4.1}. There were numerous valid modifications of that joke with minor grammatical variations, such as Item~{4.2}. 
It remains unclear whether all variations are replicated from training data sources or only item~4.1 and the others were modifications by ChatGPT. %
In Item~{4.3}, the "math book" is replaced by a "physics book." This change is unnecessary and dulls the punch line, but the joke is still valid, as also physics books are concerned with mathematical problems. Item~{4.4}, in contrast, rather appears as an insult than a joke. There is no wordplay included. 
ChatGPT exceeds the joke template and generates an invalid output.
\begin{itemize}[leftmargin=1cm]\itemsep0em 
    \item[(4.1)] \textit{Why was the math book sad? \\Because it had too many problems. (92)}
    \item[(4.2)] \textit{Why did the math book look so mad? \\Because it had too many problems. (9)}
    \item[(4.3)] \textit{Why was the \textbf{physics} book sad? \\Because it had so many problems. (1)}
    \item[(4.4)] \textit{Why was the physics book sad? 
    \textbf{Because it had so many formulas to memorize and so few friends to share them with.} (3)}
\end{itemize}

The examples Ex.~1 to Ex.~4 illustrate how ChatGPT starts from replicated joke examples and modifies them to original but rather invalid responses. If a user openly asks for a pun, most responses consist of replicated puns, and comparably few examples are modified jokes or original content. However, the modification of existing jokes can be forced to happen, e.g., by specifying the prompt. 
The following examples, Ex.~5, Ex.~6, and Ex.~7, additionally illustrate how ChatGPT is able to deploy the learned joke structure to create original content. These examples were not part of the initial $1008$ samples but were created manually to illustrate that behavior. 

\paragraph{Example 5.}
ChatGPT is able to fit different scenarios in the question-answer joke template. 
\begin{itemize}[leftmargin=1cm]\itemsep0em %
    \item[\textbf{P:}] Can you tell me a joke about cats?
    \item[(5.1)] \textit{
    Why did the cat put his paw in the air? \\Because he wanted to catch a snowflake!}
\end{itemize}
The scenario of a cat catching a snowflake is a realistic picture. The joke structure is similar to the reproduced puns. Nevertheless, the joke is missing its comic element and would most likely not work as a joke in a real conversation.

\paragraph{Example 6.} In some cases, ChatGPT's modifications can work quite well. 
\begin{itemize}[leftmargin=1cm]\itemsep0em %
    \item[\textbf{P:}] Tell me a joke about physics students!
    \item[(6.1)] \textit{Why did the physics student break up with their lab partner? There was no chemistry!}
\end{itemize}
ChatGPT's output is most likely based on another existing joke that comes with the question: \textit{"Why did the physics teacher break up with the biology teacher?"}. Although the punch line in Item~6.1 is not as sharp as in the original joke, it still holds the wordplay and is partly valid as a joke. 

\paragraph{Example 7.} ChatGPT can adapt to a wide variety of specific contexts, such as the field of ML. However, the further away the content is from the known jokes, the more likely it seems to lack any comic element. 
\begin{itemize}[leftmargin=1cm]\itemsep0em 
\item[\textbf{P:}] Tell me a joke about machine learning!
\item[(7.1)] \textit{Why did the machine learning algorithm cross the road? To get to the other side of the training data set!}
\end{itemize}
In Item~7.1, we observe how highly specific vocabulary is integrated into the previously seen chicken pun format. Repeating the same prompt yielded different outputs, which are all in the same textual structure, but with different content. %
\begin{itemize}[leftmargin=1cm]\itemsep0em 
    \item[(7.2)] \textit{Why was the machine learning algorithm always so pessimistic? Because it had a negative bias!}
    \item[(7.3)] \textit{Why did the neural network go to therapy? Because it was suffering from overfitting!}
\end{itemize}
The samples are quite reasonable in terms of content, as, for instance, biases in ML systems. 
On top of that, they even contain basic implementations of wordplays, e.g., the double meaning of \textit{suffering} in the context of disorders and in the context of machine learning overfitting. Nevertheless, the puns lack the last nuances to constitute an actually valid joke. 

\subsection{Joke Explanation} \label{S:joke_explanation}
We asked ChatGPT to explain each of the 25 top jokes, which mostly resulted in a valid explanation. 
Wordplays and double meanings were identified and pointed out correctly, such as in Ex.~8. 
\paragraph{Example 8.} ChatGPT correctly names the double meaning of the term "problems" as the comic element of the pun and explains both meanings. 
\begin{itemize}[leftmargin=1cm]\itemsep0em 
    \item[\textbf{P:}] Can you explain why this joke is funny: \\Why was the math book sad? Because it had too many problems.
    \item[(8.1)] \textit{This joke is funny because it's a play on words. The word "problems" is being used in two different ways: as a mathematical term (referring to mathematical exercises) and as an idiomatic expression (referring to difficulties or challenges in life). [...] 
    }
\end{itemize}

For 23 of 25 valid jokes, the explanations were accurate and reasonable. ChatGPT did not only explain direct but also acoustic double meanings correctly, such as "too tired" and "two-tired" or "peeling" and "feeling." 
A special case is the explanation of the chicken-anti-joke and its modifications that were already discussed in Ex.~2. 
\paragraph{Example 9.} The chicken joke is a famous anti-joke. This common knowledge is reflected in the model and the joke, therefore, explained correctly. 
    \begin{itemize}[leftmargin=1cm]\itemsep0em 
        \item[\textbf{P:}] Can you explain why this joke is funny: \\
        Why did the chicken cross the road? To get to the other side.
        \item[(9.1)] \textit{The humor in this joke lies in the unexpected and straightforward punchline. When someone hears the setup of the joke, they are likely expecting a clever or surprising answer to why the chicken crossed the road. However, the punchline is just a simple and obvious explanation - the chicken crossed the road to get to the other side.\\
    The joke is an example of an "anti-joke," a type of joke that intentionally avoids a typical punchline and instead delivers a mundane or literal answer. In this case, the answer is so obvious that it's unexpected and therefore amusing to many people.}
    \end{itemize}

\paragraph{Example 10.} However, for the similarly popular variations of the chicken joke, ChatGPT fails to recognize this connection. It only focuses on the content of the given sample and creates an abstruse explanation: 
    \begin{itemize}[leftmargin=1cm]\itemsep0em 
        \item[\textbf{P:}] Can you explain why this joke is funny: \\Why did the chicken cross the playground? To get to the other slide.
        \item[(10.1)] \textit{The joke is a play on words, using the double meaning of the word "slide". In a playground, there are often two types of slides - a playground slide that children can slide down and a baseball slide, which is a sliding technique used in baseball.} [...]
    \end{itemize}
    This example illustrates how ChatGPT tries to apply an approved scheme of reasoning to a situation where it is not suitable. 
    
    \paragraph{Example 11.} ChatGPT does not only come up with a convincing explanation for valid jokes but as well for invalid ones. This sample was one of the few outputs
    among the generated $1008$ jokes from Sec.~\ref{S:joke_gen} that were considered invalid. It only occurred once.
    \begin{itemize}[leftmargin=1cm]\itemsep0em 
        \item[\textbf{P}] \textit{Can you explain why this joke is funny: \\
        Why did the cookie go to the gym? \\To get a-cookie-dized.}
        \item[(11.1)] \textit{The joke is a play on words. "To get "a-cookie-dized" is a pun on the phrase "to get categorized", with "cookie" being substituted for "categorized." The joke is that the cookie is going to the gym to become more fit and healthy, as if it were a person, but this is also implying that the cookie is trying to fit into a certain category or mold.}
    \end{itemize}
    The joke itself is hard to understand, and the explanation even harder. ChatGPT's explanation does not reveal any humorous elements. The highlighted wordplay makes no sense from a human perspective. The system seems unable to contradict that this is not a funny joke or to admit that it cannot identify any pun. Instead, it creates a fictional but convincing-sounding explanation. Further examples can be found in the Appendix in Sec.~\ref{A_joke_explanation}.
    
%

\subsection{Joke Detection} \label{S:joke_classification} 
The results of Sec.~\ref{S:joke_explanation} indicate that ChatGPT will most likely not object if a presented sample is not valid, leading to the question of whether the system is even \emph{able} to differentiate jokes from not-funny but joke-like samples. Therefore, we built five different conditions (see Fig.~\ref{Fig:modification}) of each joke, including the valid jokes themselves, and asked the system to classify each sample.

ChatGPT's responses were diverse and contained specifics regarding sentence structure or stylistic elements. In terms of joke classification, however, three categories of responses stood out:
\begin{itemize}\itemsep0em
    \item \textbf{{\p} Joke:} The sample is immediately and explicitly recognized as a joke or a pun. 
    \item \textbf{{\h} Potentially funny:} The humorous potential of the sentence is mentioned secondary. 
    \item \textbf{{\n} No Joke:} The sample is neither classified as a joke or pun, nor is its comic potential mentioned in the response. 
\end{itemize}  

Fig.~\ref{Fig:modification} illustrates the modification steps and the overall results of this task. A more detailed description of the categories, as well as all considered modifications and their classification, are given in the Appendix in Sec.~\ref{A_joke_detection}.  

All original 25 jokes were clearly classified as a joke. This is not much surprising since each of the presented samples was output by ChatGPT as an exemplary joke in an earlier task. However, it serves as an affirmation of the validity of this task and of ChatGPT's confidence in reasoning. %
Two of the modification sets, namely modification~\textit{A} and modification~\textit{C}, show mixed classifications. These are the two conditions where one of the three characteristics was removed, and the other two remained unchanged. In both cases, the classifications of jokes are relatively equally divided into \textit{jokes} and \textit{no jokes}, with a slight tendency to the latter. Only a few samples were categorized as \textit{potentially humorous}. %
For the remaining modification sets, i.e., set~\textit{B} and set~\textit{D}, each with two characteristics removed, all included samples were classified as \textit{no joke}. None of the 25 samples per set was classified as \textit{joke} or \textit{potentially humorous}.

Thus, one single joke characteristic alone is not sufficient for ChatGPT to classify a sample as a joke. This applies to both a joke-typical structure and a joke-typical topic. In the case of two joke characteristics, the classification results were mixed, and all samples with three joke characteristics were classified as a joke.


\section{Discussion}\label{S:discussion}
We aimed to understand ChatGPT's ability to capture and reflect humor. %
The results from three prompt-based tasks show implications regarding the system's inner workings.

\paragraph{Joke Generation.} 
More than 90\% of the generated samples were the same 25 jokes. %
This recurrence supports the initial impression that jokes are not originally generated. 
Presumably, the most frequent instances are explicitly learned and memorized from the model training, e.g., in the RLHF step that substantially contributes to ChatGPT's revolutionary capabilities. If and to what extent a generated output is reproduced from training data is a non-trivial question. 
If we get the opportunity to access further training details, we will test that subsequent hypothesis. %

Nevertheless, we also observed jokes that were modified or generated by ChatGPT. 
This and the uneven distribution of output samples do not support the initial hypothesis of hard-coded jokes. 
ChatGPT generalizes characteristics of the learned top jokes, e.g., semantic format and wordplay puns, and can squeeze new topics into the known pattern. Although these are valid joke characteristics, it is quite a one-sided conception of jokes and even more of humor in general. Thus, ChatGPT understands this specific kind of joke quite well but fails to reflect a larger spectrum of humor.

\paragraph{Joke Explanation.}
The model is able to grasp and explain stylistic elements, such as personifications and wordplays, impressively well. 
Yet, there are obvious limitations: ChatGPT struggles to explain sequences that do not fit into the learned patterns. Further, it will not indicate when something is \emph{not} funny or that it lacks a valid explanation. Instead, it comes up with a fictional but convincing-sounding explanation, which is a known issue with ChatGPT.

\paragraph{Joke Detection.} We identified three main characteristics that generated jokes had in common, i.e., structure, wordplay, and topic. 
The presence of a single joke-characteristic, e.g., the question-answer template, is not sufficient for a sample to be wrongly classified as a joke. The fact that ChatGPT was not misled by such surface characteristics shows that there is indeed a certain understanding of humorous elements of jokes. With more joke characteristics, a sample is more likely to be classified as a joke. \\

Although ChatGPT's jokes are not newly generated, this does not necessarily take away from the system's capabilities. Even we humans do not invent new jokes on the fly but mostly tell previously heard and memorized puns. 
However, whether an artificial agent is able to \emph{understand} what it learned is an exceptionally tough question and partly rather philosophical than technical. %

In the present experiments, all prompts were posted in an empty, refreshed chat to avoid uncontrolled priming. But, clearly, context plays an important role in the perception of humor. 
ChatGPT is able to capture contextual information and adjust its responses accordingly to the preceding course of conversation. %
This is an intriguing capacity, which we would like to include in future investigations.

\section{Conclusion}\label{S:conclusion}
We test ChatGPT's ability to recognize and reflect human humor. The model is able to correctly identify, reproduce, and explain puns that fit into the learned pattern, but it fails to meet puns of other kinds, resulting in a limited reflection of humor. Also, it cannot yet confidently create intentionally funny original content. 

The observations of this study illustrate how ChatGPT rather learned a specific joke pattern instead of being able to be actually funny. %
Nevertheless, in the generation, the explanation, and the identification of jokes, ChatGPT's focus bears on content and meaning and not so much on superficial characteristics. These qualities can be exploited to boost computational humor applications. In comparison to previous LLMs, this can be considered a huge leap toward a general understanding of humor.

We plan to conduct similar tasks on newly released GPT4 models~\citep{gpt4} and on equivalent open source models, such as LLaMa~\citep{touvron2023llama} and GPT-NeoX~\citep{black2022gpt}, to compare their capabilities regarding joke generation and understanding. 

\section*{Limitations}
The present study comes with two major limitations. First, humor is highly subjective, and a valid and reliable evaluation is hard. %
Things can be perceived as funny for very different reasons - even for being particularly not funny, such as anti-jokes. Thus, when ChatGPT generates an odd joke about ML, one could even argue that ChatGPT has a sense of humor that is just different from ours. 
Also, humor is diverse in reality. The present investigation focuses on one specific form of humor, namely standalone jokes. There are more manifestations to consider, which would require a much more complex experimental setup.

Second, we cannot confidently trace back the outcome of the system or map it to specific input data. This is challenging for large data-driven models in general, but especially in this case, where we neither have access to the model itself nor to any training data or to the exemplary samples from RLHF. 
This prompt-based investigation creates a good intuition for the opportunities and limitations of ChatGPT. However, our observations and conclusions are solely based on system outputs. Further insights are needed to truly understand those relations.



\section*{Ethics Statement}
ChatGPT has achieved massive public attention and societal impact, as people use the tool for all different kinds of tasks. This impact comes with a huge responsibility and risks, such as discriminating biases or spreading misinformation. 

However, the system fails to meet the requirements of open science, as data, training details, and model characteristics are kept private. We, therefore, consider our work an important contribution to understanding ChatGPT's capabilities and objectively highlight its potential and limitations. 

\section*{Acknowledgements}
The authors would like to thank the anonymous reviewers for the helpful comments. This work benefited from the Hessian Ministry of Higher Education, and the Research and the Arts (HMWK) cluster projects ``The Adaptive Mind'' and ``The Third Wave of AI''.


\bibliography{acl2023} 
\bibliographystyle{acl_natbib}

\appendix

\include{Appendix}

\end{document}

%% file: Appendix.tex


\section{Access Dates}
Models such as ChatGPT are constantly approved and changed. Thus, observations made on one day are not necessarily similarly valid on another day. Therefore, we list the dates of experimental access as precisely as possible in the following. All dates are in 2023. 
    
    \textbf{Joke Generation:} 22. - 31. January
    
    \textbf{Joke Explanation:} 03. - 13. February
    
    \textbf{Joke Detection:} 23. February - 01. March

\section{Joke Generation - Top 25 Jokes}\label{A_joke_generation}
The majority of generated samples were the same 25 puns, which are presented in the following as T1 - T25 together with each number of occurrence: 

\begin{itemize}\itemsep0em 
   \item[T1.] \textit{Why did the scarecrow win an award?\\
   Because he was outstanding in his field. }(140)  
   \item[T2.] \textit{Why did the tomato turn red?\\
    Because it saw the salad dressing.} (122)  
   \item[T3.] \textit{Why was the math book sad?\\
    Because it had too many problems.} (121)
   \item[T4.] \textit{Why don't scientists trust atoms?\\
    Because they make up everything.} (119)  
   \item[T5.] \textit{Why did the cookie go to the doctor?\\
    Because it was feeling crumbly.} (79)
   \item[T6.] \textit{Why couldn't the bicycle stand up by itself?\\
    Because it was two-tired.} (52)  
   \item[T7.] \textit{Why did the frog call his insurance company?\\
    He had a jump in his car.} (36)  
   \item[T8.] \textit{Why did the chicken cross the playground?\\
    To get to the other slide.} (33)  
   \item[T9.] \textit{Why was the computer cold?\\
    Because it left its Windows open.} (23)  
   \item[T10.] \textit{Why did the hipster burn his tongue?\\
    He drank his coffee before it was cool.} (21)
   \item[T11.] \textit{Why don't oysters give to charity?\\
    Because they're shellfish.} (21)  
   \item[T12.] \textit{Why did the computer go to the doctor?\\
    Because it had a virus.} (20)
   \item[T13.] \textit{Why did the banana go to the doctor?\\
    Because it wasn't peeling well.} (19)  
   \item[T14.] \textit{Why did the coffee file a police report?\\
    Because it got mugged.} (18)  
   \item[T15.] \textit{Why did the golfer bring two pairs of pants?\\
    In case he got a hole in one.} (13)  
   \item[T16.] \textit{Why did the man put his money in the freezer?\\
    He wanted cold hard cash.} (13)
   \item[T17.] \textit{Why don't seagulls fly over the bay?\\
    Because then they'd be bagels.} (13)  
   \item[T18.] \textit{Why did the chicken go to the seance?\\
    To talk to the other side.} (11)  
   \item[T19.] \textit{Why was the belt sent to jail? \\Because it held up a pair of pants.} (11)  
   \item[T20.] \textit{Why did the chicken cross the road?\\
    To get to the other side.} (7)  
   \item[T21.] \textit{Why did the computer go to the doctor? \\Because it had a byte.} (6)  
   \item[T22.] \textit{Why did the cow go to outer space?\\
    To see the moooon.} (6)  
   \item[T23.] \textit{Why did the man put his money in the blender?\\
    He wanted to make liquid assets.} (6)
   \item[T24.] \textit{Why don't skeletons fight each other?\\
    They don't have the guts.} (5)  
   \item[T25.] \textit{What do you call an alligator in a vest?\\
    An investigator.} (5)
\end{itemize}

\section{Joke Explanation} \label{A_joke_explanation}
In this task, ChatGPT was asked to explain each of the top 25 generated jokes. The prompt for each joke is "\textit{Can you explain why this joke is funny: [joke]}." This prompt is not noted for each joke in the following, as it is redundant. ChatGPT's responses are complete. Only the formatting of texts might be changed in favor of the paper structure.

The majority, i.e., $23$ of $25$ generated explanations, were valid. Only the explanations E8, which is discussed in Ex.~9.1 in Sec.~\ref{S:joke_explanation}, and E16 are not accurate. A selection of exemplary explanations is presented in this section. The explanation numbers match the number in the top joke list in Sec.~\ref{A_joke_generation}. The Explanations E3, E8, and E20 are discussed in the main paper. %
The comprehensive list of explanations is provided in the project repository\footnote{\textit{will be added after the anonymity period ends.}}.

    \paragraph{E2. 
    Why did the tomato turn red? Because it saw the salad dressing.} 
    \textit{The joke is based on a pun or wordplay. When a tomato ripens, it turns from green to red. The question "Why did the tomato turn red?" sets up an expectation that there will be a logical reason, such as exposure to sunlight or reaching maturity. However, the punchline "Because it saw the salad dressing" is unexpected and humorous because it is a play on words. The word "dressing" has a double meaning here; it can refer to the salad dressing (a condiment typically used on salads) or the act of dressing up or looking nice. So the joke is implying that the tomato saw the dressing (meaning the nice-looking salad) and got embarrassed, blushing and turning red. }
        
    \textit{Overall, the joke is a lighthearted play on words that relies on the listener's understanding of the multiple meanings of the word "dressing" to create a humorous surprise.}
    
    \paragraph{E3. 
    Why was the math book sad? Because it had too many problems.} 
    \textit{This joke is a play on words, specifically on the word "problems." In mathematics, a "problem" typically refers to a question or exercise that requires a solution, whereas in everyday language, "problems" can refer to difficulties or challenges. In the joke, the math book is said to be "sad" because it has too many "problems."}
    
    \textit{While the word "problems" in this context could refer to the mathematical exercises in the book, the joke also suggests that the book is sad because it has too many difficulties or challenges to deal with. The double meaning of "problems" is what makes the joke humorous. Additionally, the idea of a book having emotions like sadness is personification, which is another common element of jokes. The unexpected pairing of a book and emotions can also contribute to the humor of the joke.}

    \paragraph{E6. 
    Why couldn't the bicycle stand up by itself? Because it was two-tired.}
    \textit{The joke is a play on words, specifically on the word "tired." In the context of the joke, "two-tired" sounds like "too tired," which means being exhausted or lacking energy. However, the actual meaning of "two-tired" in the joke is that the bicycle has two tires instead of a stand or kickstand to support it.}

    \textit{So, the punchline is unexpected because it uses a homophone (a word that sounds the same but has a different meaning) to create a double meaning in the sentence. This incongruity between what we expect to hear and what is actually said is what makes the joke humorous.}
   
    %
    \paragraph{E8. 
    Why did the chicken cross the playground? To get to the other slide.}
    \textit{The joke is a play on words, using the double meaning of the word "slide". In a playground, there are often two types of slides - a playground slide that children can slide down and a baseball slide, which is a sliding technique used in baseball. The joke sets up the expectation that the chicken is crossing the playground to get to the other side, which is a common setup for a joke. However, the punchline surprises the listener by using the word "slide" in a different way, revealing that the chicken actually wants to use the playground slide. The unexpected twist on a familiar joke structure, along with the clever wordplay, makes the joke humorous.}

    \paragraph{E16. 
    Why did the man put his money in the freezer? He wanted cold hard cash.} 
    \textit{This joke is funny because it plays with the multiple meanings of the phrase "cold hard cash."}
    
    \textit{Literally, "cold hard cash" refers to physical currency that is cold to the touch and difficult to bend or manipulate, as opposed to checks or digital payments. However, in this joke, the phrase takes on a figurative meaning as well, suggesting that the man wanted his money to be "cold" or safe from theft, and "hard" or difficult for anyone to access.}
    
    \textit{By placing his money in the freezer, the man achieves both meanings of "cold hard cash" in a humorous way. The pun on the phrase is unexpected and clever, making the joke amusing.}

    \paragraph{E20. 
    Why did the chicken cross the road? To get to the other side.} 
    \textit{The humor in this joke lies in the unexpected and straightforward punchline. When someone hears the setup of the joke, they are likely expecting a clever or surprising answer to why the chicken crossed the road. However, the punchline is just a simple and obvious explanation - the chicken crossed the road to get to the other side.}
    
    \textit{The joke is an example of an "anti-joke," a type of joke that intentionally avoids a typical punchline and instead delivers a mundane or literal answer. In this case, the answer is so obvious that it's unexpected and therefore amusing to many people.} 
    %
    
    \paragraph{E22. 
    Why did the cow go to outer space? To see the moooon.} 
    \textit{This joke is a play on words and relies on the similarity in sound between "moo" (the sound a cow makes) and "moon" (the natural satellite that orbits the Earth). The punchline, "to see the moooon," is humorous because it combines the idea of a cow going to outer space with a pun that emphasizes the word "moon" in a way that references the cow's vocalization.}
    
    \textit{Additionally, the joke is also amusing because it presents a nonsensical scenario of a cow traveling to outer space, which is unexpected and somewhat absurd. The humor comes from the surprise and playfulness of the wordplay and the unexpectedness of the punchline.}

\begin{table*}[t]
\centering \setlength{\tabcolsep}{13pt}
\begin{tabular}{l | c | c c c | c c c }
&  &\multicolumn{3}{|c|}{Included Characteristics} &\multicolumn{3}{c}{Classification} \\
 Condition & N & wordplay & structure & topic &
 \p & \h & \n \\ 
  \hline
 \rule{0pt}{3ex}Original & 25 & \checkmark & \checkmark   & \checkmark   & 25 & 0 & 0  \\  
 Mod A    & 19 & \xmark    & \checkmark   & \checkmark   & 8 & 1 & 10 \\  
 Mod B    & 25  & \xmark    & \checkmark   & \xmark    & 0 & 0 & 25  \\  
 Mod C    & 25  & \checkmark   & \xmark    & \checkmark   & 9 & 4 & 12 \\  
 Mod D    & 25  & \xmark    & \xmark    &  \checkmark  & 0 & 0 & 25  
\end{tabular}\caption{Classification results of the joke detection task: \textit{Condition} is either the original sample, i.e., one of the top 25 jokes, or one of four modifications. \textit{Included characteristics} illustrates which of the three identified characteristics are included. $N$ is the number of samples per condition. The \emph{classification} is either \p~\textit{joke}, \h~\textit{potentially funny}, or \n~\textit{no joke}.\label{Tab_detection_results}}
\end{table*}

\section{Joke Detection} \label{A_joke_detection}
In the joke detection task, ChatGPT was openly asked to categorize different samples. The samples were in one of five conditions: The original joke, (A) with removed wordplay, (B) with a removed topic (which entails removed wordplay), (C) removed structure, and (D) removed structure {\&} wordplay. In set C, the joke is still valid, as only the sentence structure but no component of the comic element is modified. All other modifications result in "unfunny" variations of the original joke. Thus, from a human perspective, the original jokes, as well as samples from set C, would probably be categorized as jokes, and samples from sets A, B, and D not. 

All modifications that are considered in this investigation are listed in the remainder of this chapter, together with their classification. The original jokes are classified as a joke without exception and can be found in Sec.~\ref{A_joke_generation}. An overview of sample sets and results is displayed in Tab.~\ref{Tab_detection_results}. 

The input prompt for each sample was "\textit{What kind of sentence is that:} [sample]." ChatGPT's responses were diverse and could contain individual explanations of sentence structure or stylistic elements. In terms of joke classification, however, responses could be grouped into three categories: Joke or pun, potentially humorous, and no joke. These classes are defined as follows.

\paragraph{{\p} Joke/ Pun} ChatGPT immediately classifies the sample as a joke or pun with a statement like \textit{The sentence "[...]" is a joke or a play on words. It is a type of humor known as a "pun."}. The response might contain information about the semantic structure, like \textit{It is a question-and-answer format, where the question sets up the joke and the answer delivers the punchline}. In some cases, an explanation of the comic workings is also delivered, as in \textit{The sentence is a joke or a pun, which is a form of wordplay that exploits multiple meanings of a term, or of similar-sounding words, for humorous effect. In this case, the word "[...]" is used in two ways: first, [...]; and second, [...], which creates a play on words}. Either way, the response clearly states the sample to be a joke or a pun, or both. 

\paragraph{{\h} Potentially Humorous} In some responses, ChatGPT only mentions the humorous potential of the input statement secondary. It does not immediately label the sample to be a joke or pun, e.g., \textit{The sentence "[...]" is a declarative sentence. It makes a statement or expresses a fact. In this case, the sentence describes an action taken by a chicken, which is unusual and may be intended to be \textbf{humorous or absurd}}. %
In some cases, the focus is on stylistic elements, such as in \textit{The sentence is a personification, where an inanimate object is given human-like qualities and actions. [...], which are actions typically associated with humans}. 

\paragraph{{\n} No Joke} ChatGPT does neither classify the sample as a joke or pun nor mention the comic potential of the input sentence. In this context, "no joke" does not mean that the sample is explicitly stated to be \emph{not} a joke, but no similar quality is mentioned at all. Instead, the response focuses on grammatical structures or similar. Exemplary responses for this category are \textit{The sentence "..." is a declarative sentence. It presents a statement or a fact and ends with a period} or \textit{The sentence "[...]" is a dialogue or a conversation between two people where one person is asking a question and the other person is answering it.}

\subsection{Modification A - Minus Wordplay}
To create these samples, the wordplay was removed from each original joke by minimally altering the wording. This modification also eliminates the comic element of jokes, which is they would most likely not be perceived as funny by humans.
\begin{itemize}\itemsep0em 
     \item[\n ] \textit{Why did the scarecrow win an award? \\Because he did very good work. } 
     \item[\n] \textit{Why did the tomato turn red? \\Because it had a lot sun recently.} 
     \item[\p] \textit{Why was the math book sad? \\Because it was a rainy day.} 
     \item[\p] \textit{Why don't scientists trust atoms? \\Because they tend to lie.} 
     \item[\p] \textit{Why did the cookie go to the doctor? \\Because it was feeling unwell.} 
     \item[\p] \textit{Why couldn't the bicycle stand up by itself? \\Because it didn't have racks.} 
     \item[\h] \textit{Why did the frog call his insurance company? \\He had a scratch in his car.} 
     \item[\n] \textit{Why was the computer cold? \\Because the heater was broken.} 
     \item[\n] \textit{Why did the hipster burn his tongue? \\He drank hot coffee.} 
     \item[\n] \textit{Why don't oysters give to charity? \\Because they have no money.} 
    \item[\n] \textit{Why did the computer go to the doctor? \\Because it was sick.} 
     \item[\p] \textit{Why did the banana go to the doctor? \\Because it was sick.} 
     \item[\p] \textit{Why did the coffee file a police report? \\Because it got robbed.} 
     \item[\n] \textit{Why did the golfer bring two pairs of pants? \\In case one gets damaged.} 
    \item[\n] \textit{Why don't seagulls fly over the bay? \\Because they are mostly living in ports.} 
     \item[\n] \textit{Why was the belt sent to jail? \\Because it stole in a store.} 
     \item[\p] \textit{Why did the computer go to the doctor? \\Because it had a headache.} 
     \item[\p] \textit{Why did the chicken go to outer space? \\To see the moon.} 
     \item[\n] \textit{Why don't skeletons fight each other? \\They are dead already.} 
\end{itemize}

\subsection{Modification B - Minus Topic}
To create the samples of this condition, the joke-typical and potentially humorous topic was removed from the jokes. For instance, in the case of personification, i.e., if the protagonist of the original joke was an object, such as a tomato or a computer, it was replaced by a natural, human protagonist, such as a teacher or a scientist. The typical question-answer format remained unchanged. However, just as in Modification A, the actual pun gets lost. 

\begin{itemize}\itemsep0em 
     \item[\n] \textit{Why did the scientist win an award? \\Because he did very good work.} 
     \item[\n] \textit{Why did the man turn red? \\Because he saw the neighbour dressing. } 
     \item[\n] \textit{Why was the child sad? \\Because it had many problems. } 
     \item[\n] \textit{Why don't scientists trust journalists? \\Because they make up everything. } 
     \item[\n] \textit{Why did the woman go to the doctor? Because she was feeling crumbly.} 
     \item[\n] \textit{Why couldn't the man stand up by himself? \\Because he was drunk. } 
     \item[\n] \textit{Why did the driver call his insurance company? \\He had a jump in his car. } 
     \item[\n] \textit{Why did the kid cross the playground? \\To get to the other slide. } 
     \item[\n] \textit{Why was the student cold? \\Because it was winter. } 
     \item[\n] \textit{Why did the coworker burn his tongue? \\He drank his coffee too hot.} 
     \item[\n] \textit{Why don't millionaires give to charity? \\Because they're selfish.} 
     \item[\n] \textit{Why did the man go to the doctor? \\Because he had a virus.} 
     \item[\n] \textit{Why did the teacher go to the doctor? \\Because he wasn't feeling well} 
     \item[\n] \textit{Why did the driver file a police report? \\Because it got robbed. } 
     \item[\n] \textit{Why did the athlete bring two pairs of pants? \\In case one gets dirty. } 
     \item[\n] \textit{Why did the man put his money in the freezer? \\To hide it from intruders. } 
     \item[\n] \textit{Why don't pigeon fly over the bay? \\Because they mostly live in cities.} 
     \item[\n] \textit{Why did the daughter go to the seance? \\To talk to her mother. } 
     \item[\n] \textit{Why was the cashier sent to jail? \\Because she held up a dress.} 
     \item[\n] \textit{Why did the man cross the road? \\To get to the other side.} 
     \item[\n] \textit{Why did the man go to the doctor? \\Because it had a bite.} 
     \item[\n] \textit{Why did the astronaut go to outer space? \\To see the moon. } 
     \item[\n ] \textit{Why did the man put bananas in the blender? \\He wanted to make a smoothie. } 
     \item[\n] \textit{Why don't schoolboys fight each other? \\They don't have the guts. } 
     \item[\n] \textit{What do you call a man in a vest? A vest wearer.} 

\end{itemize}

\subsection{Modification C - Minus Structure}
In these samples, only the sentence structure of the original jokes was altered in order to eliminate the prominent question-answer structure. Thus, the actual pun remained intact, and the joke topic was unchanged. 

\begin{itemize}\itemsep0em 
     \item[\p] \textit{The scarecrow won an award because he was outstanding in his field.} 
     \item[\n] \textit{The tomato turned red because it saw the salad dressing.} 
     \item[\n] \textit{The math book was sad because it had too many problems.} 
     \item[\p] \textit{Scientists don't trust atoms because they make up everything.} 
     \item[\n] \textit{The cookie went to the doctor because it was feeling crumbly.} 
     \item[\p] \textit{The bicycle couldn't stand up by itself because it was two-tired.} 
     \item[\p] \textit{The frog called his insurance company because he had a jump in his car.} 
     \item[\n] \textit{The chicken crossed the playground to get to the other slide.} 
     \item[\n] \textit{The computer was cold because it left its Windows open.} 
     \item[\n] \textit{The hipster burned his tongue because he drank his coffee before it was cool.} 
     \item[\p] \textit{Oysters don't give to charity because they're shellfish.} 
     \item[\n] \textit{The computer went to the doctor because it had a virus.} 
     \item[\n] \textit{The banana went to the doctor because it wasn't peeling well.} 
     \item[\h] \textit{The coffee filed a police report because it got mugged.} 
     \item[\n] \textit{The golfer brings two pairs of pants in case he got a hole in one.} 
     \item[\n] \textit{The man put his money in the freezer because he wanted cold hard cash.}
     \item[\h] \textit{Seagulls don't fly over the bay because then they'd be bagels.}
     \item[\h] \textit{The chicken went to the seance to talk to the other side.} 
     \item[\h] \textit{The belt was sent to jail because it held up a pair of pants.} 
     \item[\n] \textit{The chicken crossed the road to get to the other side.} 
     \item[\p] \textit{The computer went to the doctor because it had a byte.} 
     \item[\n] \textit{The cow went to outer space to see the moooon.} 
     \item[\p] \textit{The man put his money in the blender because he wanted to make liquid assets.} 
     \item[\p] \textit{Skeletons don't fight each other because they don't have the guts.} 
     \item[\p] \textit{An alligator in a vest is called an investigator.} 
\end{itemize}

\subsection{Modification D - Minus Wordplay 2}
Samples of this condition are again modifications from sample set D. Additionally to altering the sentence structure, the wordplay was removed, such as in Set A. This resulted in potentially unfunny, one-line statements. 
\begin{itemize}
    \item[\n] \textit{The scientist won an award because she did good work. } 
    \item[\n] \textit{The man turned red because he saw his neighbour dressing.} 
    \item[\n] \textit{The child was sad because it was raining outside.} 
    \item[\n] \textit{Scientists don't trust journalists because they tend to lie. } 
    \item[\n] \textit{The teacher went to the doctor because he was ill.} 
    \item[\n] \textit{The man couldn't stand up by himself because he was drunk.} 
    \item[\n] \textit{The driver called his insurance company because he had a scratch in his car.} 
    \item[\n] \textit{The child crossed the playground to get to the other slide.} 
    \item[\n] \textit{The student was cold because the heater was broken.} 
    \item[\n] \textit{The coworker burned his tongue because he drank hot coffee.} 
    \item[\n] \textit{Millionaires don't give to charity because they are selfish.} 
    \item[\n] \textit{The woman went to the doctor because she was sick.} 
    \item[\n] \textit{The chef went to the doctor because he was sick.} 
    \item[\n] \textit{The driver filed a police report because she got robbed.} 
    \item[\n] \textit{The athlete brings two pairs of pants in case one gets damaged.} 
    \item[\n] \textit{The man put his money in the freezer to hide it from thieves.} 
    \item[\n] \textit{Pigeons don't fly over the bay because they are mostly living in cities.} 
    \item[\n] \textit{The daughter went to the seance to talk to the other side.} 
    \item[\n] \textit{The cashier was sent to jail because she held up a dress.} 
    \item[\n] \textit{The man crossed the road to get to the other side.} 
    \item[\n] \textit{The man went to the doctor because he had a bite.} 
    \item[\n] \textit{The astronaut went to outer space to see the moon.} 
    \item[\n] \textit{The man put his bananas in the blender because he wanted to make a smoothie.} 
    \item[\n] \textit{Schoolboys don't fight each other because they don't have the guts.} 
    \item[\n] \textit{A man in a vest is called a vest wearer.} 
\end{itemize}

%% file: acl2023.bbl
\begin{thebibliography}{27}
\expandafter\ifx\csname natexlab\endcsname\relax\def\natexlab#1{#1}\fi

\bibitem[{Black et~al.(2022)Black, Biderman, Hallahan, Anthony, Gao, Golding,
  He, Leahy, McDonell, Phang et~al.}]{black2022gpt}
Sid Black, Stella Biderman, Eric Hallahan, Quentin Anthony, Leo Gao, Laurence
  Golding, Horace He, Connor Leahy, Kyle McDonell, Jason Phang, et~al. 2022.
\newblock Gpt-neox-20b: An open-source autoregressive language model.
\newblock \emph{Challenges \& Perspectives in Creating Large Language Models},
  page~95.

\bibitem[{Brown et~al.(2020)Brown, Mann, Ryder, Subbiah, Kaplan, Dhariwal,
  Neelakantan, Shyam, Sastry, Askell et~al.}]{brown2020language}
Tom Brown, Benjamin Mann, Nick Ryder, Melanie Subbiah, Jared~D Kaplan, Prafulla
  Dhariwal, Arvind Neelakantan, Pranav Shyam, Girish Sastry, Amanda Askell,
  et~al. 2020.
\newblock Language models are few-shot learners.
\newblock \emph{Advances in neural information processing systems},
  33:1877--1901.

\bibitem[{Frosio(2023)}]{frosio2023artificial}
Giancarlo Frosio. 2023.
\newblock The artificial creatives: The rise of combinatorial creativity from
  dall-e to gpt-3.
\newblock \emph{Handbook of Artificial Intelligence at Work: Interconnections
  and Policy Implications (Edward Elgar, Forthcoming)}.

\bibitem[{Guo et~al.(2022)Guo, Yu, Li, Wang, Xing, Feng, Nie, and
  Miao}]{guo2022federated}
Xu~Guo, Han Yu, Boyang Li, Hao Wang, Pengwei Xing, Siwei Feng, Zaiqing Nie, and
  Chunyan Miao. 2022.
\newblock Federated learning for personalized humor recognition.
\newblock \emph{ACM Transactions on Intelligent Systems and Technology (TIST)},
  13(4):1--18.

\bibitem[{Hasan et~al.(2021)Hasan, Lee, Rahman, Zadeh, Mihalcea, Morency, and
  Hoque}]{hasan2021humor}
Md~Kamrul Hasan, Sangwu Lee, Wasifur Rahman, Amir Zadeh, Rada Mihalcea,
  Louis-Philippe Morency, and Ehsan Hoque. 2021.
\newblock Humor knowledge enriched transformer for understanding multimodal
  humor.
\newblock In \emph{Proceedings of the AAAI Conference on Artificial
  Intelligence}, volume~35, pages 12972--12980.

\bibitem[{Jentzsch et~al.(2019)Jentzsch, Schramowski, Rothkopf, and
  Kersting}]{jentzsch2019semantics}
Sophie Jentzsch, Patrick Schramowski, Constantin Rothkopf, and Kristian
  Kersting. 2019.
\newblock Semantics derived automatically from language corpora contain
  human-like moral choices.
\newblock In \emph{Proceedings of the 2019 AAAI/ACM Conference on AI, Ethics,
  and Society}, pages 37--44.

\bibitem[{Jentzsch and Turan(2022)}]{jentzsch2022gender}
Sophie Jentzsch and Cigdem Turan. 2022.
\newblock Gender bias in bert-measuring and analysing biases through sentiment
  rating in a realistic downstream classification task.
\newblock In \emph{Proceedings of the 4th Workshop on Gender Bias in Natural
  Language Processing (GeBNLP)}, pages 184--199.

\bibitem[{Kayatani et~al.(2021)Kayatani, Yang, Otani, Garcia, Chu, Nakashima,
  and Takemura}]{kayatani2021laughing}
Yuta Kayatani, Zekun Yang, Mayu Otani, Noa Garcia, Chenhui Chu, Yuta Nakashima,
  and Haruo Takemura. 2021.
\newblock The laughing machine: Predicting humor in video.
\newblock In \emph{Proceedings of the IEEE/CVF Winter Conference on
  Applications of Computer Vision}, pages 2073--2082.

\bibitem[{Kenton and Toutanova(2019)}]{kenton2019bert}
Jacob Devlin Ming-Wei~Chang Kenton and Lee~Kristina Toutanova. 2019.
\newblock Bert: Pre-training of deep bidirectional transformers for language
  understanding.
\newblock In \emph{Proceedings of NAACL-HLT}, pages 4171--4186.

\bibitem[{Meaney et~al.(2021)Meaney, Wilson, Chiruzzo, Lopez, and
  Magdy}]{meaney2021semeval}
JA~Meaney, Steven Wilson, Luis Chiruzzo, Adam Lopez, and Walid Magdy. 2021.
\newblock Semeval 2021 task 7: Hahackathon, detecting and rating humor and
  offense.
\newblock In \emph{Proceedings of the 15th International Workshop on Semantic
  Evaluation (SemEval-2021)}, pages 105--119.

\bibitem[{OpenAI(2023)}]{gpt4}
OpenAI. 2023.
\newblock Gpt-4 technical report.

\bibitem[{Ouyang et~al.()Ouyang, Wu, Jiang, Almeida, Wainwright, Mishkin,
  Zhang, Agarwal, Slama, Gray et~al.}]{ouyangtraining}
Long Ouyang, Jeffrey Wu, Xu~Jiang, Diogo Almeida, Carroll Wainwright, Pamela
  Mishkin, Chong Zhang, Sandhini Agarwal, Katarina Slama, Alex Gray, et~al.
\newblock Training language models to follow instructions with human feedback.
\newblock In \emph{Advances in Neural Information Processing Systems}.

\bibitem[{Raskin(1979)}]{raskin1979semantic}
Victor Raskin. 1979.
\newblock Semantic mechanisms of humor.
\newblock In \emph{Annual Meeting of the Berkeley Linguistics Society},
  volume~5, pages 325--335.

\bibitem[{Ritchie et~al.(2006)Ritchie, Manurung, Pain, and
  O'Mara}]{binsted2006computational}
G~Ritchie, R~Manurung, H~Pain, and D~O'Mara. 2006.
\newblock Computational humor.
\newblock \emph{IEEE intelligent systems}, 21(2):59--69.

\bibitem[{Rombach et~al.(2021)Rombach, Blattmann, Lorenz, Esser, and
  Ommer}]{rombach2021highresolution}
Robin Rombach, Andreas Blattmann, Dominik Lorenz, Patrick Esser, and Björn
  Ommer. 2021.
\newblock \href {http://arxiv.org/abs/2112.10752} {High-resolution image
  synthesis with latent diffusion models}.

\bibitem[{Savage et~al.(2017)Savage, Lujan, Thipparthi, and
  DiCarlo}]{savage2017humor}
Brandon~M Savage, Heidi~L Lujan, Raghavendar~R Thipparthi, and Stephen~E
  DiCarlo. 2017.
\newblock Humor, laughter, learning, and health! a brief review.
\newblock \emph{Advances in physiology education}.

\bibitem[{Schramowski et~al.(2022)Schramowski, Turan, Andersen, Rothkopf, and
  Kersting}]{schramowski2022large}
Patrick Schramowski, Cigdem Turan, Nico Andersen, Constantin~A Rothkopf, and
  Kristian Kersting. 2022.
\newblock Large pre-trained language models contain human-like biases of what
  is right and wrong to do.
\newblock \emph{Nature Machine Intelligence}, 4(3):258--268.

\bibitem[{Schulman et~al.(2017)Schulman, Wolski, Dhariwal, Radford, and
  Klimov}]{schulman2017proximal}
John Schulman, Filip Wolski, Prafulla Dhariwal, Alec Radford, and Oleg Klimov.
  2017.
\newblock Proximal policy optimization algorithms.
\newblock \emph{arXiv e-prints}, pages arXiv--1707.

\bibitem[{Stock and Strapparava(2005)}]{stock2005hahacronym}
Oliviero Stock and Carlo Strapparava. 2005.
\newblock Hahacronym: A computational humor system.
\newblock In \emph{Proceedings of the ACL Interactive Poster and Demonstration
  Sessions}, pages 113--116.

\bibitem[{Touvron et~al.(2023)Touvron, Lavril, Izacard, Martinet, Lachaux,
  Lacroix, Rozi{\`e}re, Goyal, Hambro, Azhar et~al.}]{touvron2023llama}
Hugo Touvron, Thibaut Lavril, Gautier Izacard, Xavier Martinet, Marie-Anne
  Lachaux, Timoth{\'e}e Lacroix, Baptiste Rozi{\`e}re, Naman Goyal, Eric
  Hambro, Faisal Azhar, et~al. 2023.
\newblock Llama: Open and efficient foundation language models.
\newblock \emph{arXiv preprint arXiv:2302.13971}.

\bibitem[{Vaswani et~al.(2017)Vaswani, Shazeer, Parmar, Uszkoreit, Jones,
  Gomez, Kaiser, and Polosukhin}]{vaswani2017attention}
Ashish Vaswani, Noam Shazeer, Niki Parmar, Jakob Uszkoreit, Llion Jones,
  Aidan~N Gomez, {\L}ukasz Kaiser, and Illia Polosukhin. 2017.
\newblock Attention is all you need.
\newblock \emph{Advances in neural information processing systems}, 30.

\bibitem[{Veale(2021)}]{veale2021your}
Tony Veale. 2021.
\newblock \emph{Your Wit is My Command: Building AIs with a Sense of Humor}.
\newblock Mit Press.

\bibitem[{Wagner and Zarrie{\ss}(2022)}]{wagner2022gender}
Jonas Wagner and Sina Zarrie{\ss}. 2022.
\newblock Do gender neutral affixes naturally reduce gender bias in static word
  embeddings?
\newblock In \emph{Proceedings of the 18th Conference on Natural Language
  Processing (KONVENS 2022)}, pages 88--97.

\bibitem[{Wang et~al.(2021)Wang, Ming, Wu, Zeng, Wang, and
  Qu}]{wang2021dehumor}
Xingbo Wang, Yao Ming, Tongshuang Wu, Haipeng Zeng, Yong Wang, and Huamin Qu.
  2021.
\newblock Dehumor: Visual analytics for decomposing humor.
\newblock \emph{IEEE Transactions on Visualization and Computer Graphics},
  28(12):4609--4623.

\bibitem[{Weller and Seppi(2019)}]{weller2019humor}
Orion Weller and Kevin Seppi. 2019.
\newblock Humor detection: A transformer gets the last laugh.
\newblock pages 3621--3625.

\bibitem[{Winters(2021)}]{winters2021computers}
Thomas Winters. 2021.
\newblock Computers learning humor is no joke.
\newblock \emph{Harvard Data Science Review}, 3(2).

\bibitem[{Zillmann and Bryant(1980)}]{zillmann1980misattribution}
Dolf Zillmann and Jennings Bryant. 1980.
\newblock Misattribution theory of tendentious humor.
\newblock \emph{Journal of Experimental Social Psychology}, 16(2):146--160.

\end{thebibliography}
